\documentclass[annual,pagebackref=true]{acmsiggraph}

\usepackage{times}

\def\mytitle{Real-Time Character Rise Motions}

\title{\fontsize{18pt}{18pt}\selectfont\mytitle}

\author{Ben Kenwright\thanks{e-mail:bkenwright@ieee.org (First Drafted Dec 2013)}}

\pdfauthor{Ben Kenwright}

\def\mykeywords{Rising, Biped, Animation, Inverse Kinematics, Falling, Balancing, Physically-Based, Standing-Up, Getting-Up}

\keywords{\mykeywords}

\graphicspath{{./images/}}

\newcommand{\figuremacroW}[4]{
	\begin{figure} 
		\centering
		\includegraphics[width=#4\columnwidth]{#1}
		\caption[#2]{\textbf{#2} - #3}
		\label{fig:#1}
	\end{figure}
}

\newcommand{\figuremacroF}[4]{
	\begin{figure*} 
		\centering
		\includegraphics[width=#4\textwidth]{#1}
		\caption[#2]{\textbf{#2} - #3}
		\label{fig:#1}
	\end{figure*}
}

\usepackage{amsmath}

\usepackage{enumitem}

\hypersetup{
    colorlinks = true, 
    linkcolor = blue,
    anchorcolor = red,
    citecolor = blue, 
    filecolor = red, 
}

\hypersetup{pdfinfo={
   Author		= {Ben Kenwright},
   Title		= {\mytitle},
   Subject 		= {\mytitle},
   CreationDate = {D:20130530195600},
   Keywords 	= {\mykeywords}},
   pdfcreator   = {Ben Kenwright},
   pdfproducer  = {Ben Kenwright}
}

\begin{document}


\teaser{
   \includegraphics[width=1.0\textwidth]{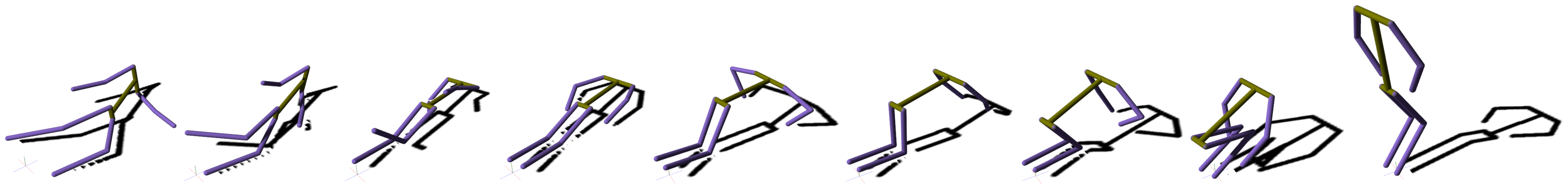}
   \caption{Rise from Laying.  The rise controller model could be mapped onto skeletons of different levels of complexity to reduce ambiguity and singularity issues.  The figure shows the full-body skeleton represented as a basic 9-link stick man rolling over and getting-up from the front.}
}

\maketitle

\begin{abstract}
	This paper presents an uncomplicated dynamic controller for generating physically-plausible three-dimensional full-body biped character rise motions on-the-fly at run-time.  Our low-dimensional controller uses fundamental reference information (e.g., center-of-mass, hands, and feet locations) to produce balanced biped get-up poses by means of a real-time physically-based simulation.  The key idea is to use a simple approximate model (i.e., similar to the inverted-pendulum stepping model) to create continuous reference trajectories that can be seamlessly tracked by an articulated biped character to create balanced rise-motions.  Our approach does not use any key-framed data or any computationally expensive processing (e.g., offline-optimization or search algorithms).  We demonstrate the effectiveness and ease of our technique through example (i.e., a biped character picking itself up from different laying positions).
\end{abstract}


\begin{CRcatlist} 
		\CRcat{I.3.7}{Computer Graphics}{Three-Dimensional Graphics and Realism}{Animation}
		\CRcat{I.6.8}{Computer Graphics}{Animation}{Types of Simulation--Animation}
\end{CRcatlist}

\keywordlist


\copyrightspace


\section{Introduction} \label{sec:introduction}

\noindent
\textbf{Motivation:} 
While a tremendous amount of research has been made over the past decade on controllers and animations that have focused specifically on virtual characters' upright motions (e.g., standing, walking, and dancing) \cite{YLP07,Ken12,KM99,Ken11,CBV10}.  Less research has addressed the issue of how a character would regain its balance by picking itself-up (e.g., after falling down).  We focus on a biped rise controller that does not depend on any key-framed motion capture libraries, is simple, easy-to-implement, and remarkably robust; since, an algorithmic approach has the potential of producing a more general solution which is capable of generating unique physically-plausible movements.

\noindent
\textbf{Interest $\&$ Importance:} 
We want our animated characters to appear as realistic and life-like as possible.  Hence, our characters should be able to fall-down (e.g., due to disturbances from pushes or trips) and mimic the real-world.  Therefore, it would be significant if the characters could generate adaptable, physically-plausible, and natural-looking motions to synthesize the characters picking themselves-up.  A physics-based approach allows us to create motions that are interactive, dynamic, and customizable, while possessing physically correct-properties to produce visually plausible life-like rise animations; for example, when a biped rises from a laying pose it needs to shift its center-of-mass and move its hands and feet while maintaining its balance to reach its final goal of an upright standing posture.  Finally, if we use the physical properties of a character (e.g., feature sizes and mass), we can customize and adapt the get-up motions without constantly needing to search for and edit pre-canned animation libraries to fit the specific situation changes.

\noindent
\textbf{Challenges:} 
Humans possess a large number of degrees-of-freedom (DOF), and it is difficult to determine joint angles that will achieve the multiple goals with varying priority and in real-time.  In addition, there are the physical attributes, whereby the character needs to use its body to maintain balance while picking itself-up.  The key challenges of our approach is generating robust poses in real-time without key-framed data that can be used to animate a full-body, three-dimensional biped to faithfully reproduce a natural realistic rise motion.  The motion need to account for ground contacts, swing-hand placement, and the center-of-mass support area balance priority considerations, while ensuring the poses are always physically-plausible and life-like.

\noindent
\textbf{Existing Solutions:} 
A popular uncomplicated method for representing character rise motion is to switch to a pre-created key-framed sequence that will make the character get-up.  The animations can possess life-like properties and provide a repertoire of unique and diverse animation solutions \cite{KGP02}.  These animation sequences can further be adapted by means of kinematic techniques, for example, so the hands and feet engage the environment (i.e., in contact with the terrain) \cite{BCPP08,LWBP*10}.  For example, a set of key poses are guided by inverse kinematics to move the character along a realistic get-up motion.  However, the kinematic motions might not always be physically-plausible, and a library of animations is needed, while the final motions cannot be easily adapted to different character features (e.g., short, tall, fat) or unique environmental situations (e.g., un-even terrain).  However, physics-based models have been used to create balanced key-pose by analyzing, planning and solving full-body problems to create rise motions \cite{BJ11,YKH04}.

\noindent
\textbf{Our Approach:} 
This paper presents a practical, straightforward, and robust system for producing physically plausible get-up motions on-the-fly at run-time.  Our model can be applied to different character feature sizes (e.g., short, tall, fat), and can produce various unique get-up motions by changing control parameters.  We focus on time critical systems, such as games, for producing practical fundamental motions \textit{without key-framed data}.  The resulting get-up motions are physically-accurate, and the generated poses are based on an uncomplicated approximation model that provides crucial balanced reference trajectory information, which can be used to control a full-body articulated biped character.  We use inverse kinematic (IK) techniques to combine our simplified base-controller with an articulated skeleton to produce fluid, physically accurate, and controllable get-up movements.

\noindent
\textbf{Contribution:} 
The key contribution of this paper is the introduction of a novel controller method for generating physically-plausible rising (i.e., get-up) character motions for real-time environments.  In summary, the main contributions of this paper are:
\begin{itemize}[noitemsep,topsep=0pt,parsep=0pt,partopsep=0pt]
	\item Real-time approach for generating rising (get-up) motions without key-framed data
	\item Low-dimensional physics-based model to produce character animations that are self-driven (i.e., the character picks itself-up) while obeying geometric and kinematic constraints (e.g., joint and reach limits) and physical laws (e.g., gravity, non-slip contacts)
	\item We demonstrate and explain our approaches simple, practical, and straightforward ability for correcting and generating balanced rise poses
\end{itemize}


\section{Related Work}
\label{sec:related_work}

There has been a broad range of exciting and interesting approaches across different disciplines (i.e., graphics, robotics, and biomechanics) towards creating and adapting character animation solutions.  Whereby, we briefly review some of the most recent and relevant research within these fields that has contributed towards synthesizing biped characters picking themselves up.

\subsection{Computer Graphics}
\label{sec:computer_graphics}

Kinematic methods use pre-generated motions, either from being painstakingly created by artists or through motion capture recordings.  Whereby, the motions can be blended together using motion graphs to achieve a particular movement \cite{KGP02}.  The motion graphs can be extended to generate \textit{paramaterized} motions that are more flexible and less repetitive by blending kinematic motions with a physically-based systems (e.g., including collision responses) \cite{BCPP08,LWBP*10}.  However, these generated motions depend on the available motion libraries to accomplish the specific action.

Faloutsos et al. \cite{FPT01} illustrated an application of their approach on rising from a supine position.  They focused on combining controllers for different types of motion.  The generated rise motion were based on a fixed posture with no mention of generating rising motions for different start poses.  The controllers used pose tracking and timed state transitions, making it difficult to transfer controllers to new characters or environments.  Essentially, using key-framed data to implement the character pushing itself up onto all four, then rising to its feet in a final upright balanced pose.

Liu et al. \cite{LYPS10} proposed a sampling-based approach to reconstruct controlled movement from underlying reference motions.  Their technique demonstrates excellent robustness while preserving physical correctness on contact-rich motions, including rolling, get-up, and kip-up.  This sampling-based approach can produce small motion variations that can be treated as noise.  However, reference motions are still needed for producing larger motion variations.  Their method uses an expensive offline process, and does not incorporate feedback into the generated controllers.

Lin et al. \cite{WCYH12}, recently demonstrated life-like rising animations by using motion planning based on an RRT (rapidly-explored random tree) based approach (i.e., picking the most plausible motion from a planned motion path).  This approach was also applied to manipulation planning by Yamane et al. \cite{YKH04}.  Whereby, the motion planning was used to compute the path of an object being manipulated.  For each planned object orientation and position, the pose of the character is computed to satisfy geometric, kinematic, and posture constraints.  While both approaches are similar, Yamane et al. \cite{YKH04} focused on object space planning, while Lin et al. \cite{WCYH12} used a posture space planning with the RRT-blossom algorithm.

Zordan et al. \cite{ZMCF05} connected a physically simulated movement to a MOCAP motion, with a focused on generating dynamic responses by tracking a desired trajectory, which is formed by linearly interpolating the intermediate postures from the two motion capture sequences before and after the transition.  Their approach synthesizes a trajectory using a posture database.  In particular, an arbitrary laying posture or key-pose can be used, but the linear interpolation approach always produces the same trajectories.  Similarly, Wrotek et al. \cite{WJM06} exploited a data-driven solution to control a physically accurate model, one of the solutions, including a rising model.

Jones \cite{BJ11} generated key-poses using a physics-based model to produce rising motions, while Nunes et al. \cite{NVC07} controlled an articulated structure using a state machine logic to accomplish show jumping, running, rising motions.  

A novel alternative was the ``motion doodle'' system presented by Thorne et al. \cite{TBP04}, which let the user sketch the intended motion path to obtain a desired representation of the movement.  This user feedback approach could synthesize appropriate motions that were visually correct; some of the example motions included, jumping and getting-up.

\subsection{Robotics} 
\label{sec:robotics}

Morimoto and Doya \cite{MMDK04,KKFH*03} proposed a hierarchical reinforcement learning method to generate standing-up movement on a simplified character.

Hirukawa et al. \cite{HKKK*05}, and Fujiwara et al. \cite{KKFH*03} divided a rising motion into several contact states and used a contact-state graph to represent them.  This approach works well on robots, but is difficult to define a proper contact-state graph for human motions rising from various laying postures.

Kanehiro et al. \cite{KFHN*07} generated getting up motions by linearly interpolating any given laying posture to its most similar posture in a predefined falling state graph on a HRP-2P robot, which is 1.5m tall and weighs 58kg.  A similar controller for a smaller robot 0.5m, 2.3kg has also been developed \cite{SSB06}.  Their work focuses on generating a smooth sequence of rise postures and less on the physical plausibility of the rising motion.

Kuniyoshi et al. \cite{KOTN*04} used an adult sized humanoid robot to analyze the critical aspects of a highly dynamic getting up motion, and used this information to find the parameters to generate successful motions.  However, transferring the motion from simulation to the robot was challenging and error prone, due to the dependence of the motion on the difficult to simulate ground contact forces.

Mettin et al. \cite{MLFS08} created rise motions for a character sitting (i.e., a chair sitting-to-standing and vice-versa motion) while keeping balanced, using torques and arm forces.  Fujiwara et al. \cite{KKFH*03} created a humanoid robot with features the same as a human that could lie down and pick itself-up.  Kuniyoshi et al. \cite{KOTN*04} examined roll-and-rise motion capture data to generate robot movements based on temporal localization of features to extract crucial information about the task.

\subsection{Biomechanics}
\label{sec:biomechanics}

Standing up motions are well studied in biomechanics, in particular, is the analysis of sit-to-stand motions, including the dynamics and stability which have been studied in detail \cite{RPMG*96}.

Muscle activity through the motion can be divided into three distinct phases: a forward lean of the upper body, an upward acceleration due to leg extension, and a deceleration phase \cite{HTO99,KJM90,RDHJ*94}.  In addition, the contact forces between the buttocks and the feet are controlled by muscle activations to generate the forward and upward acceleration to stand.

McCoy and VanSant \cite{MJVA93}, and Ford-Smith and VanSant \cite{FCVA93} compared movement patterns of people rising from a bed in different ages.  For adolescents, they developed four categories of movement patterns: far upper extremity, near upper extremity, axial region and lower extremities.  In the age between 30 to 59, they developed four categories of movement patterns: left upper limb movement patterns, right upper limb movement patterns, head and trunk movement patterns and lower limb movement patterns.  They experimented and computed the probability of each movement pattern.  The majority of the biomechanics studies are to analyze rather than generate the rising motion.

\section{System Overview}
\label{sec:system_overview}
In our approach, we use a low-dimensional model (i.e., a particle-mass with weightless telescopic arms and legs) for estimating key information that is common with a complex biped character structure to create a fast, robust, and simple solution for generating get-up motions.  The final motions enable the character to pick itself up using its feet-hand placement and by shifting its own body position to maintain balance and achieve an upright standing pose.  The controller gives initial information on where to place the character's hand and pelvis.  Then as the controller iteratively proceeds to stand up, a feedback loop between the low-dimensional controller and the character's model corrects for errors as the character slowly gets up.  The systems key focus is a character's motion that is anatomically-correct (i.e., bound to joint limits), physically-plausible (i.e., obeys mechanical laws, such as balancing), realistic motion (i.e., analogous to a real-world human), and which is computationally fast, simple, and robust.  The character picks \textbf{himself} up by positioning his body in a balanced pose and using his own joint torques.

\figuremacroW
{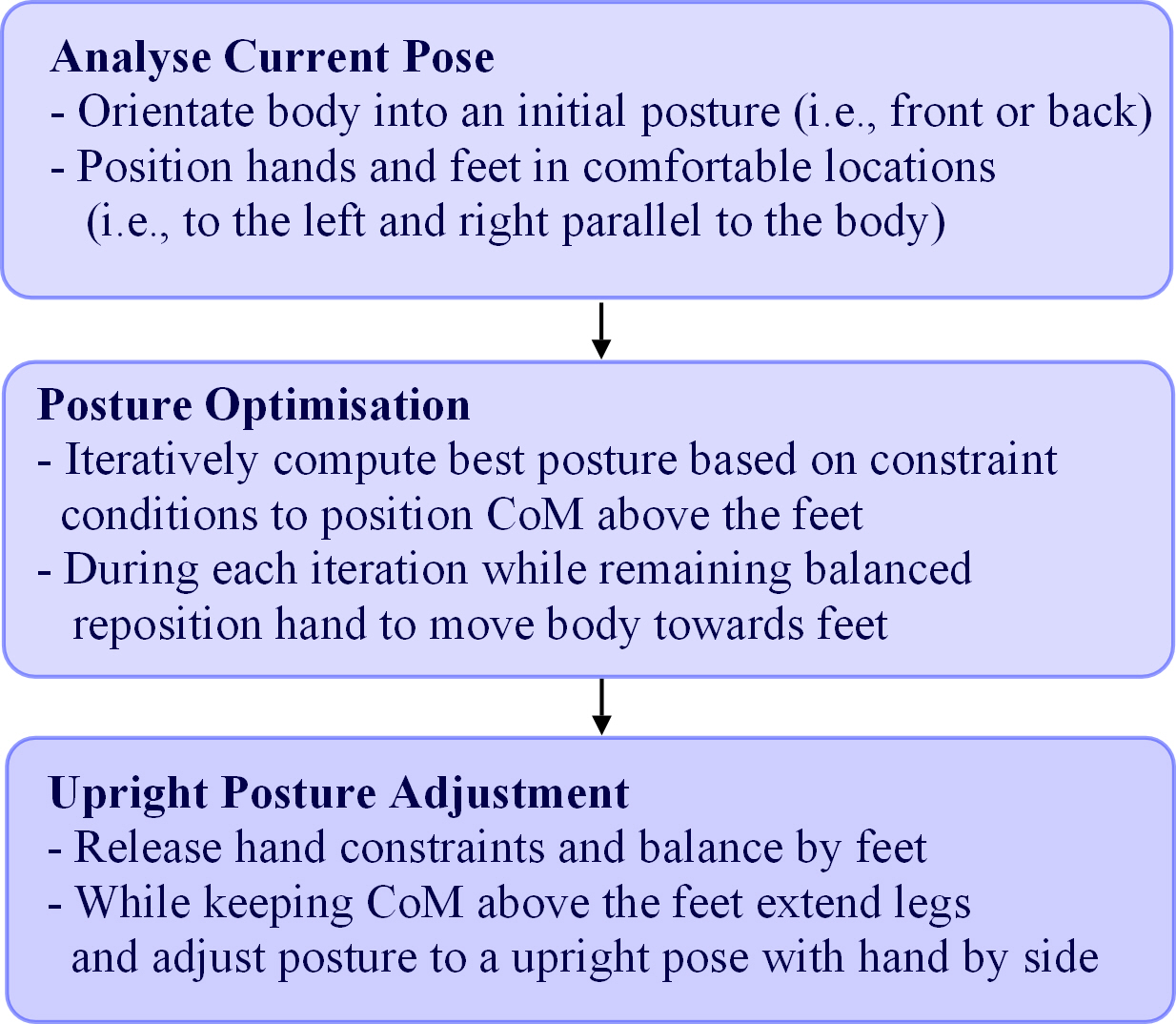}
{System Overview}
{An overview of the rising (get-up) motion framework.}
{0.9}

\section{Optimization} 
This section explains the optimization problem for generating approximate poses that accomplish the goal of getting up from laying down.  These poses need to enforce imposed constraints (e.g., CoM above the support area) to accomplish the primary task of maintaining continuous balance.  Furthermore, the poses should be as natural-looking and as comfortable as possible (i.e., avoiding contorted or overstretched positions).  We approach the problem by subdividing the task into numerous stages.  Each stage contributes towards the final goal of a vertically upright balanced posture.

\figuremacroW
{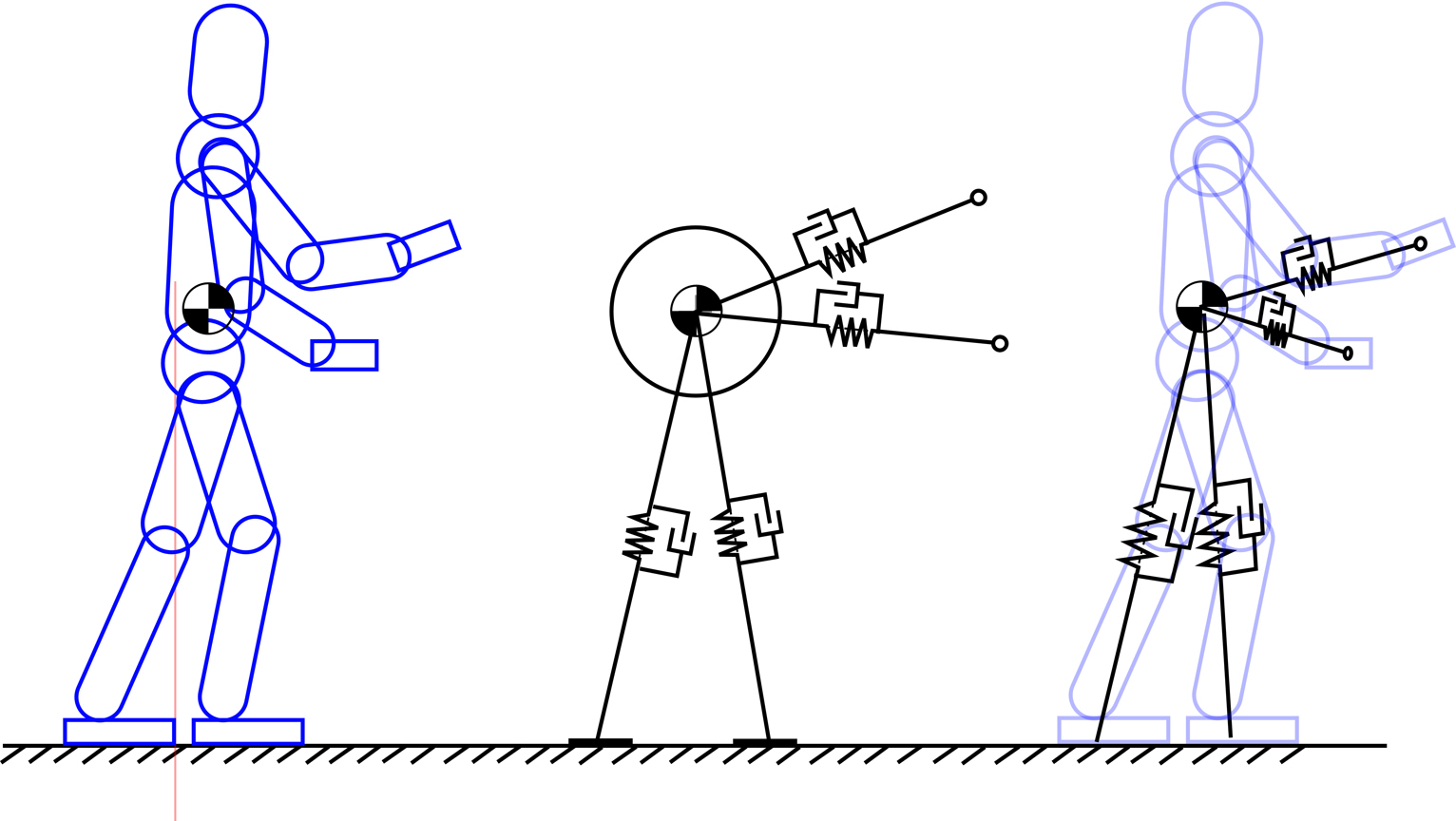}
{Simplified Point-Mass Biped Model}
{The arms and legs of the character are analogous with a spring-damper mechanism and work in synergy to control the overall character's center-of-mass (CoM).}
{0.8}

\figuremacroW
{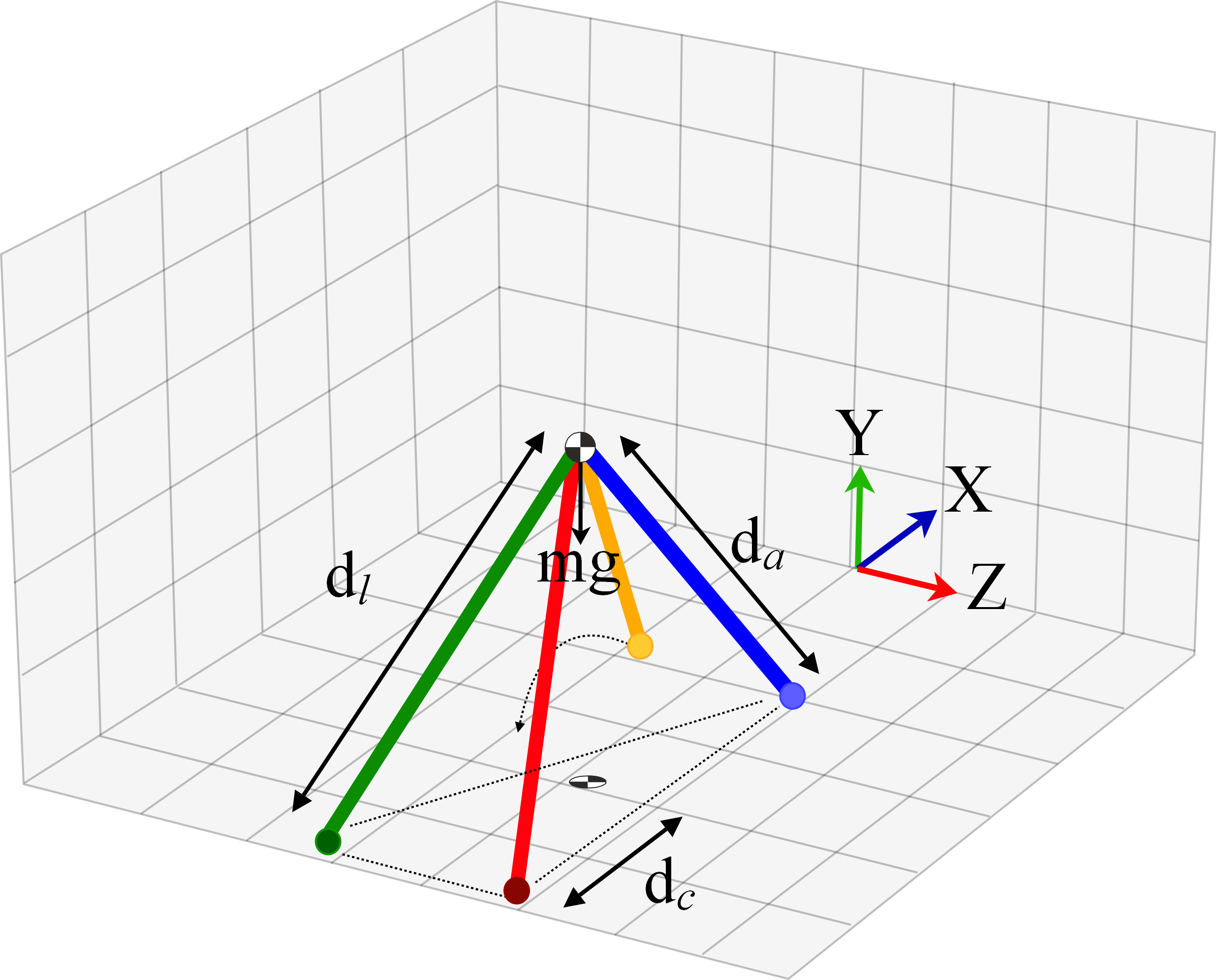}
{Simple Optimization Problem}
{We reduce the complexity of the problem down to a simple point-mass with three-contact points.  We iteratively move the CoM so that it is as horizontally close as possible to the feet (i.e., above the feet).  Then we move the free-hand closer and make it the new support hand, and repeat the process.  Figure shows the simple optimization problem.}
{0.8}

\subsection{Simplified Model}
We use a simplified model for the optimization problem.  The problem and model focus on essential elements (such as, feet, hands, CoM position, and support region).  The simplified model consists of a point-mass (m) representing the character's overall center-of-mass (CoM), two mass-less legs, and two mass-less arms (as shown in Figure \ref{fig:simple_model}).  The legs, arms, and CoM have constraints imposed upon them (e.g., minimum and maximum lengths, target location) that we find an optimum solution at each iteration to accomplish the goal of maintaining balance while moving towards an upright pose.  Hence, we evaluate the contact points (i.e., end-effectors) for each hand and foot by means of an uncomplicated systematic analysis of the balancing situation.  The simplified model enables us to reduce the complexity of the problem and find an optimized solution in real-time while retaining crucial characteristics of a character's rise motion.  For example, to demonstrate the principal driving logic for our approach, imagine a simple 2D couple-rod, shown in Figure \ref{fig:rod_get_up}, picking itself up by iteratively moving its center-of-mass above the foot support region, then it can raise its upper rod while keeping the CoM above the support region to accomplish a vertical upright pose.  Simplifying the knees and elbows to extendable rigid links with minimum and maximum lengths helps reduce ambiguity and singularities when searching for an optimum solution to the low-dimensional constraint problem. 

\figuremacroW
{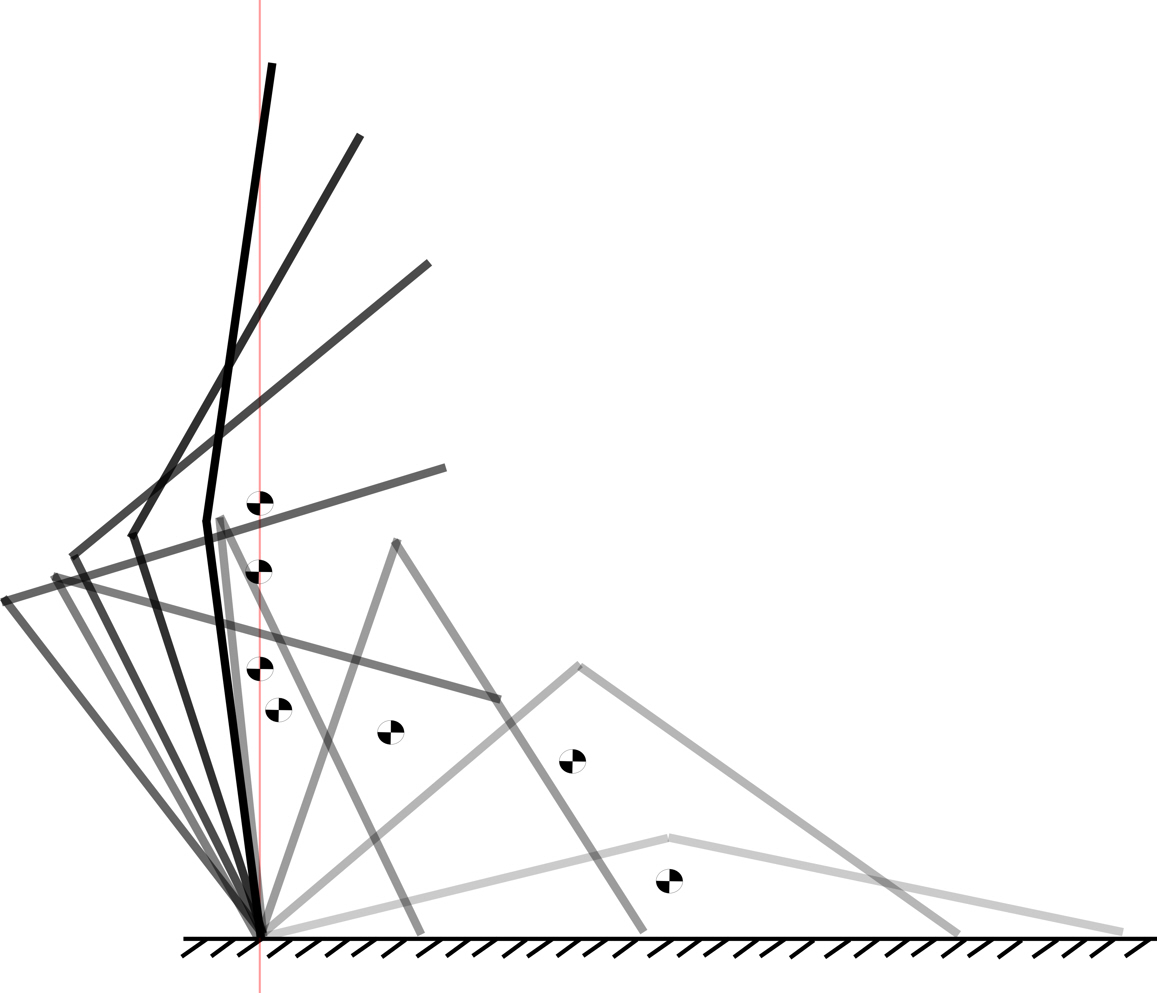}
{Rigid Rod Picking Itself Up}
{A simple rod picking itself up (assuming the bottom of the rod is shorter or of equal in length to the top - and the mass is not biased towards the top).}
{1.0}

\subsection{Geometric and kinematic optimization steps}
The model has to account for the geometric relationship between the character and the virtual environment as well as the kinematic control; for example, the coupled control between the simplified low-dimensional model and the high-dimensional articulated biped character.

\figuremacroW
{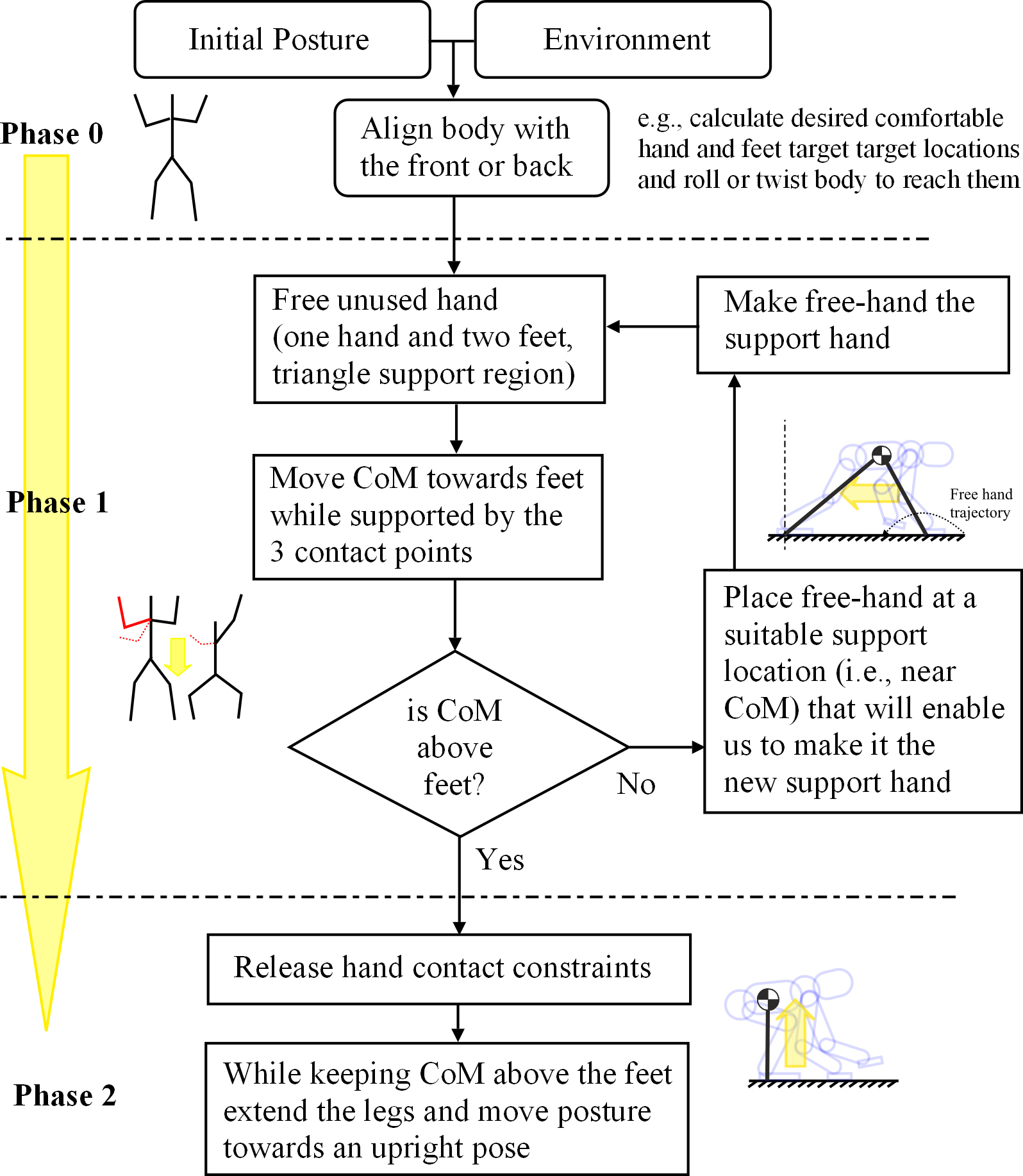}
{Flow Graph}
{Our approach uses three phases: initial pose, iteratively shifting balance to feet, and rising to an upright posture.}
{1.05}

Our model's logic is divided into three phases (as shown in Figure \ref{fig:flow_graph}).

\textbf{Phase 0 (move towards start pose)}\\
After the character is laying down, we want to have him roll onto his front or side.  So we locate comfortable hand and foot positions (i.e., four targets) to the left and right of the character's body (i.e., avoid crossed arms and legs).\\
We can roll the character and move their arms and legs towards their target start locations.  We have now started with both feet and hands' locked in place ready for the next phase.

\textbf{Phase 1 (move CoM towards the feet support region)}\\
We release one hand so the body is balanced above the two feet and single support hand.  The two feet and the support hand form a projected triangle on the ground known as the body support region.  The CoM should be within this body support region for the body remain balanced (i.e., not fall over).\\
We then search for an optimal solution for the simple model (i.e., two feet and a single hand).  We set constraints, leg and arm minimum and maximum lengths, and the CoM as close to the feet support region as possible.  \\
When we find a solution we interpolate the model towards its optimal goal (i.e., legs and arm lengths).  \\
Is the CoM above the foot support region?\\
	No: Keeping the support hand locked, we move the free hand to the side of the body at the location of the CoM (e.g., the left free hand to the left side of the body).  The free hand's location is now made the support hand, and a new body support region is formed (restart Phase 1).\\
	Yes: Balanced by our feet (goto Phase 2).\\
	
\textbf{Phase 2 (move to an upright pose)}
Release both hand constraints, since we are supported by our feet support region, as the CoM is located within it (from Iteration 1).\\
Move vertically upwards while keeping the CoM above the feet support region area.\\

The constraint optimization conditions for phases 1 and 2 are (with primary and secondary priority for importance):
\begin{equation}
\begin{alignedat}{3}
\text{Phase 1:}&
\left\{
	\begin{array}{ll}
		d_{lmin} > d_l > d_{lmax} & (primary)\\
		d_{amin} > d_a > d_{amax} & (primary)\\
		d_c = 0 				  & (secondary)\\
	\end{array}
\right. \\
\text{Phase 2:}&
	\left\{
		\begin{array}{ll}
			d_c = 0 			  & \quad\quad\quad\quad\;\; (primary)\\
			d_l = d_{lmax} 		  & \quad\quad\quad\quad\;\; (secondary)\\
		\end{array}
	\right. \\
\end{alignedat}
\end{equation}

where $d_l$ and $d_a$ are the arm and leg distance from the CoM, $d_c$ is the CoM horizontal distance from the foot support region, and subscript min and max dictate minimum maximum constraint conditions (as shown in Figure \ref{fig:logic_details}).  Foot-wise balanced poses, such as Figure \ref{fig:sphere_get_up} and Figure \ref{fig:cube_get_up}, are considered part of phase 2.  Since they are only concerned with keeping the CoM above the feet support area. 

We concentrate on slow (i.e., static) get-up motions whereby the CoM and body support area can be used to classify the balance stability criteria (i.e., for static or slow motions the zero moment point (ZMP) is equal to the projected CoM \cite{KM99}).

\figuremacroW
{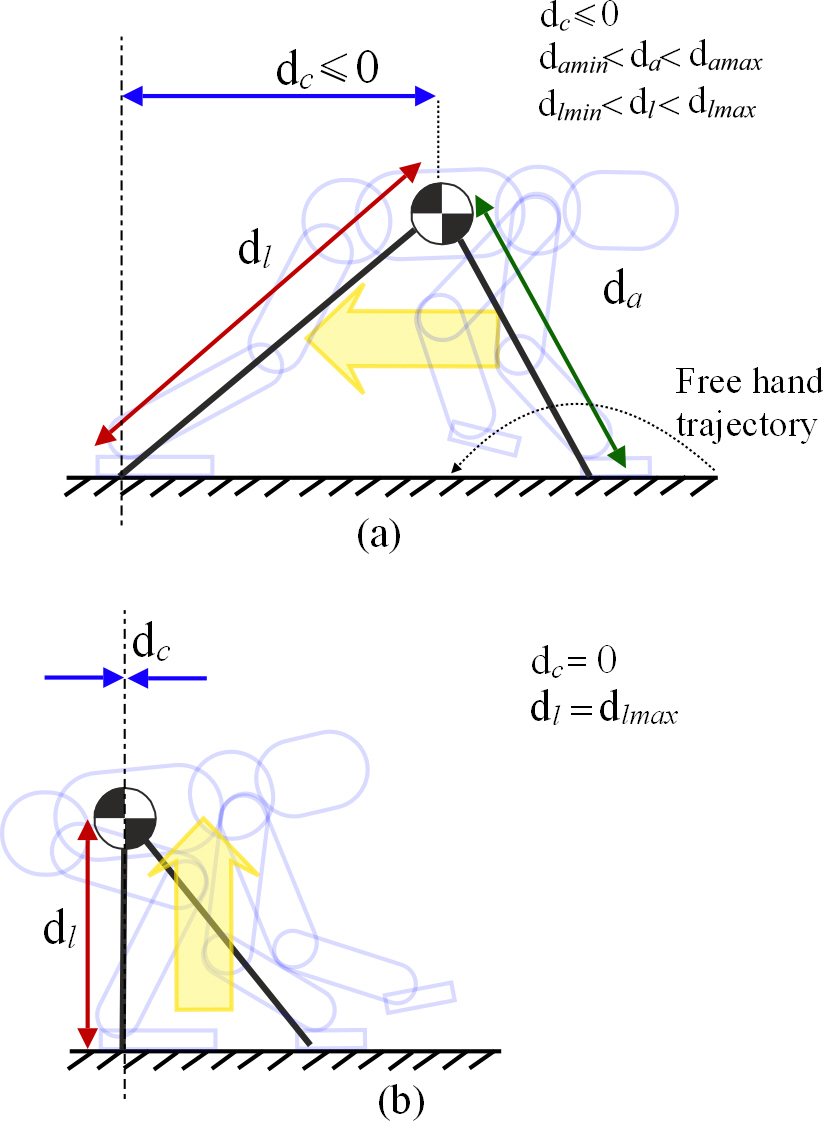}
{Rising Phases}
{(a) Phase 1, which iteratively swaps hand locations to keep moving the CoM towards the feet support region, (b) Phase 2 starts when the CoM is above the feet support region; whereby, the arm constraints are released, and the focus is on extending the legs and keeping the CoM above the feet, and is complete when the legs are fully extended and the arms come to rest at the body's side.}
{0.8}

\figuremacroW
{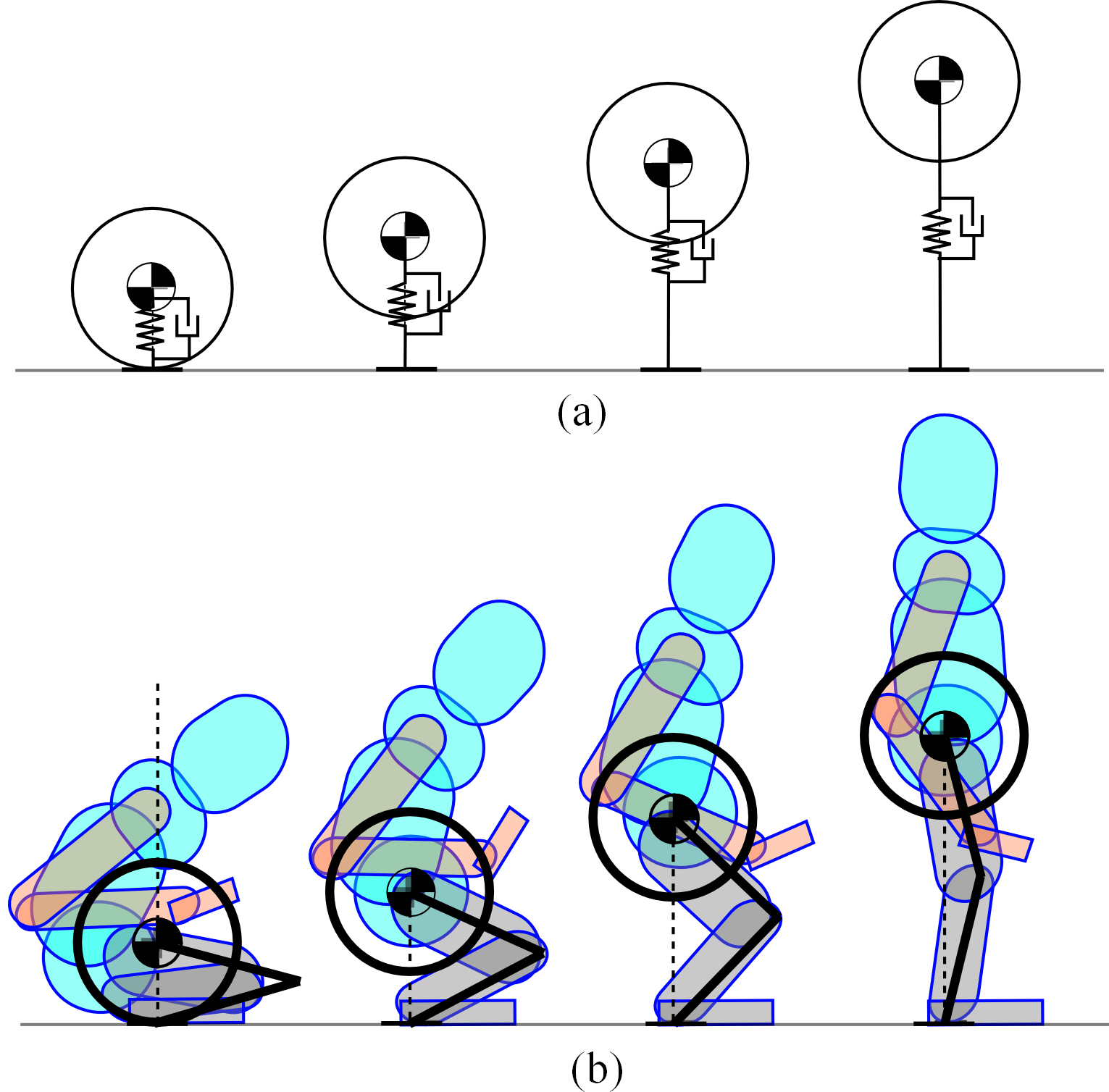}
{Arm-less Getting-Up}
{The character performs a crouch-to-rise, if the character's CoM is above the feet support area.  We can approximate the mass as a particle-point and the problem reduces to a spherical object extending the support leg length.  (a) The leg muscle is analogous to a spring-damper system extending its rest length, and (b) illustrating a crouched character on the ground.}
{0.8}

\figuremacroW
{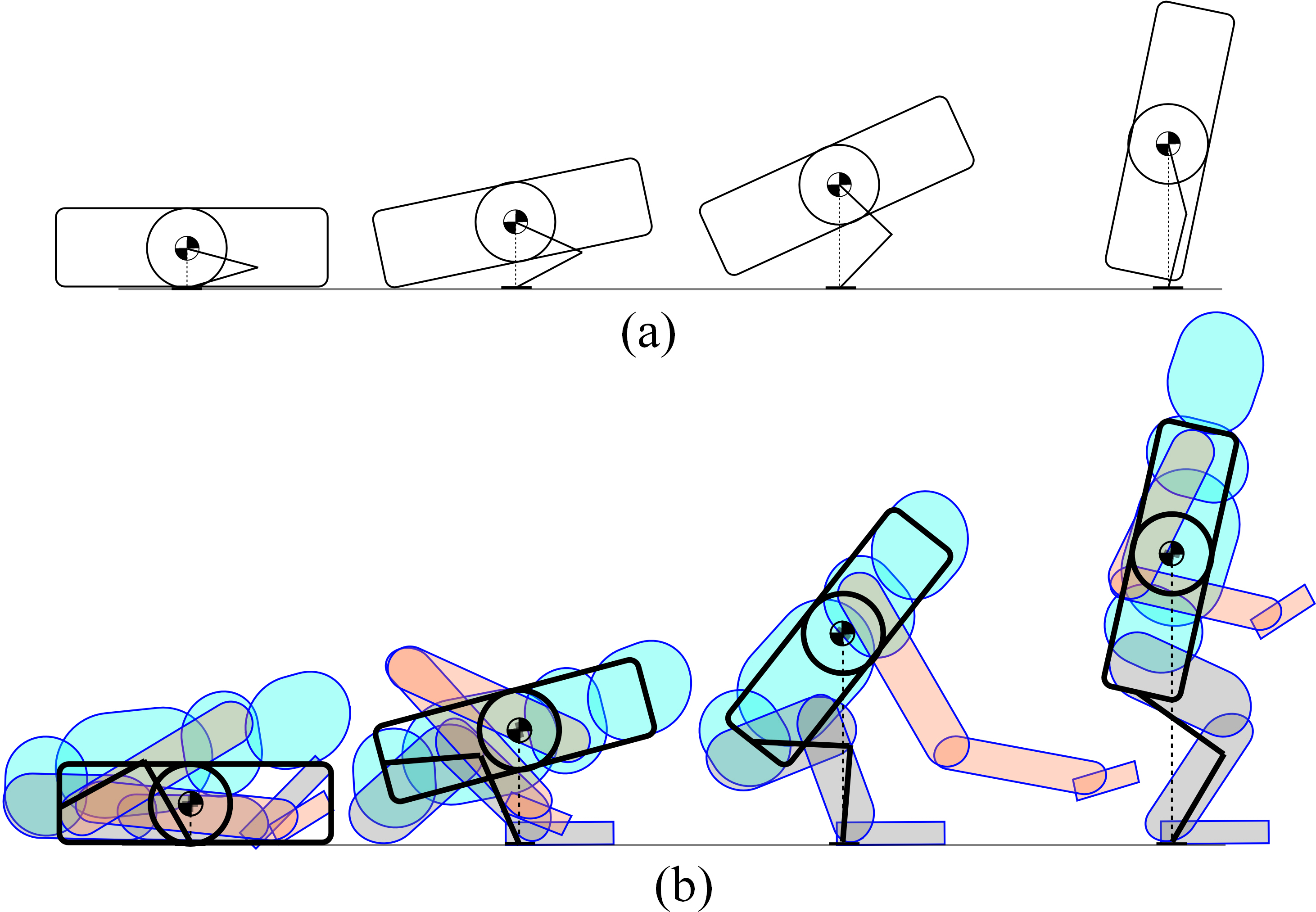}
{Arm-less Elongated Body Getting-Up}
{The uncomfortable and uncommon pose of a character stretched out can keep their CoM above their feet support area while getting-up; (a) the CoM stays above the feet support area while the posture rotates, and (b) the legs are connected to the end of the elongated body while gradually rotated and keeping the CoM above the feet and reaching the vertical stance pose.}
{1.0}

Figure \ref{fig:rod_get_up} illustrates how our controller would go about picking itself up.  Initially, it positions the center-of-mass above the center-of-pressure (CoP), from there on, it lifts its front body up, while compensating with the lower body to maintain the CoM above the foot position.  Due to the dynamic feedback from the model, any disturbances which might arise during rising, will be fed back into the base which will attempt to compensate for them.

\subsection{Trajectories (Smooth Interpolated Motions)}
The trajectories for the hands and feet are calculated using B\'ezier spline paths when moving them between old and new position during the get-up sequences (the height they are lifted above the ground and the speed they move all affect the final style of the motion).

\subsection{Controller Constraints (Priority)}
For the base controller to achieve an upright posture, a number of constraints must be imposed.  
It must be possible for the controller to place its center-of-mass above its foot position.  If both the upper and lower body have the same radial dimensions, this means the upper body must great in length than the lower body, but less than twice the length of the lower body.

The steps the controller goes through while picking itself up, are:
\begin{enumerate}[noitemsep,topsep=0pt,parsep=0pt,partopsep=0pt]
	\item Align the center-of-mass above the foot position.
	\item Slowly raise the upper body, and while doing so, compensate for center-of-mass moving outside the foot position region using the lower body.
\end{enumerate}

\subsection{Biped Model}
For our character simulation tests, we used a variety of different models.  Primarily, for 3D simulations, we used a 15-link biped model, shown in Figure \ref{fig:biped_model} and Figure \ref{fig:model}.  The character model possesses 36 degrees-of-freedom (DOF), including 6 for the world root (3 translation and 3 rotation), 1 DOF for each knee and elbow, and 2 DOF for each hip and shoulder.

\figuremacroW
{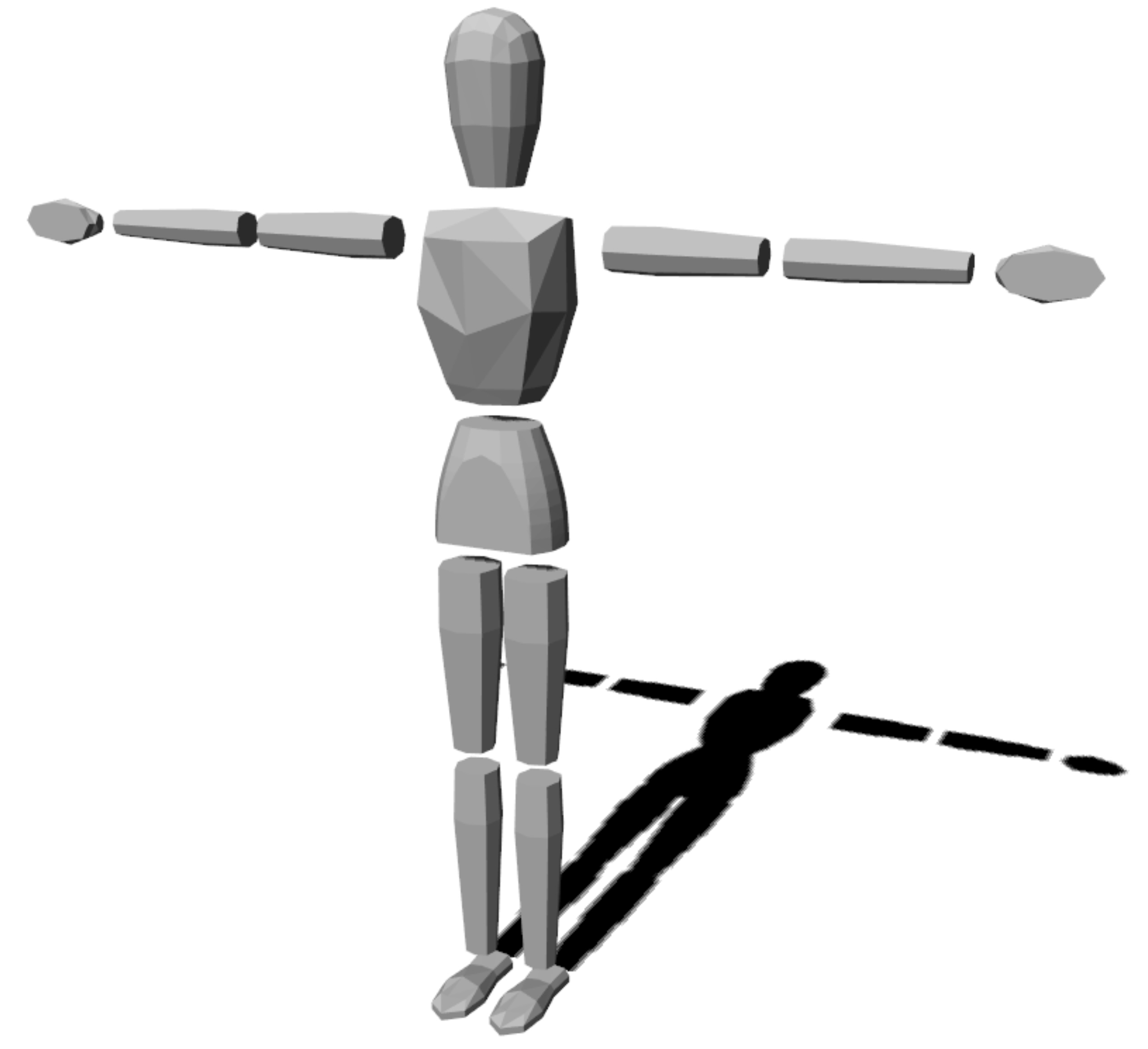}
{Articulated Biped Character Model}
{The 3D biped simulation model is composed of 15-links and 14-joints.}
{0.7}

\section{Inverse Kinematics (IK)}

We use inverse kinematics (IK) to map our low-dimensional model's information onto our articulated biped character skeleton.  The IK is responsible for generating the final biped joint angles, it also imposes physical joint angle constraints to ensure the model always produces physically-plausible poses.  Since we are interested in real-time applications, we employ a fast and simple analytic solution (as done by Coros et al. \cite{CBV10}) for two-link inverse kinematic (IK) problems given by Kulpa et al. \cite{KMA05}.  We compute the unique solution for \textbf{the elbow and knee by forcing them to lie on a plane} (e.g., the elbow plane would encumber the shoulder and hand while the knee would encumber the hip and ankle).  The rotational degree-of-freedom (DOF) of this embedded plane is specified as an input parameter that allows for bow-legged styles and control over the expressive nature of arm reaching movements.

Mapping the low-dimensional model's center-of-mass (CoM) position onto the full-body biped skeleton, there are two fundamental inverse kinematic (IK) approaches.  They are:

\begin{enumerate}[noitemsep,topsep=0pt,parsep=0pt,partopsep=0pt]
	\item A computationally fast less accurate approach - e.g., the hip midpoint as the CoM position similar to SIMBICON ~\cite{YLP07} due to it being fast and simple
	\item A more precise globally solution (CoM of all limbs) - e.g., constantly update and track the whole articulated body CoM position synonymous with the approach by Tsai et al. \cite{TLCL+10}.
\end{enumerate}

For 2D cases, we use a global CoM tracking constraint IK solution \cite{Ken12b}, while for 3D simulations, we opt for the simpler and computationally faster hip midpoint for the CoM position.  For example, the mapping of the model onto a 3D biped skeleton is shown in \ref{fig:simple_3d_ik}.  As for the hand orientation, we chose to have the hand initially rotate and align to face comfortably forwards at the start of the get-up motion and neglect any twisting.

\figuremacroW
{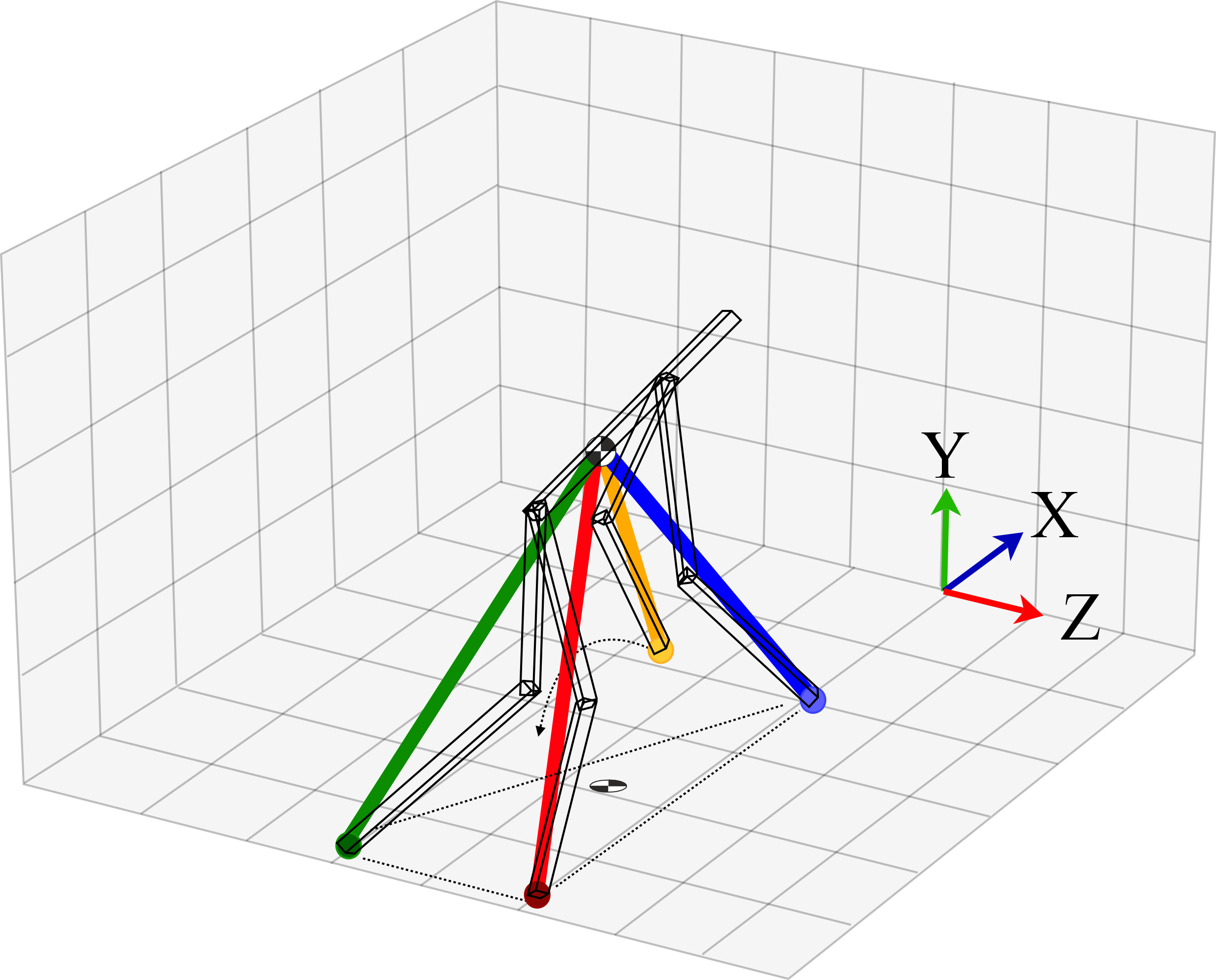}
{Mapping Simple Model onto Biped Structure}
{The inverse kinematic problem to reconstruct the biped character pose.  For example, in the figure, we use a 9 link biped model, with 18 degree-of-freedom (DOF), 6-DOF root (i.e., position and orientation), 2-DOF each hip and shoulder, 1-DOF for each knee and elbow.  It has three fixed ground contact points, one for each foot, and one for the support hand.}
{0.8}

\subsection{Style (Priority ordered IK)}
Incorporating a primary and secondary IK solver allows us to mix in behavioral motions.  The second optional constraint condition embeds optional characteristic motions while the crucially primary constraints are enforced to ensure the motion is physically-correct and balanced.  For example, we could mix in a tired sluggish movement, looking-around motions, or coherent random life-like movements (e.g., swaying and looking around) to make the movement less robot-like and unique.  This is done in two parts, using the primary key elements to keep the character balanced and physically correct (locked with the optimized model's rise solution), and a secondary motion added on top to introduce stylistic control (as shown by \cite{Ken12b}).

\section{Rigid Body Control}
The IK solver provides joint angles that we used to calculate joint torques to control the full-body rigid body skeleton structure.  This approach is analogous to a \textit{puppet on strings}, since the rigid body structure emulates the IK solution through angular springs (i.e., proportional derivative servos).  However, since the final motions are generated using an articulated rigid body structure, the movements are smoother while still possessing their responsive and interactive properties.  The joint torques for the articulated character are generated using a proportional derivative (PD) controller, i.e., $\tau  = {k_p}({\theta _d} - \theta ) - {k_d}\theta '$, where ${\theta _d}$ is the desired joint angle, $\theta$ and $\theta '$ are the current joint angle and joint angular velocity, and ${k_p}$ and ${k_d}$ are the gain and damping constants.  While the gain and damping constants are crucial for the character's motions to appear responsive and fluid, calculating reliable, robust coefficients that result in the pose reaching its desired target within a specific time reliably and safely is difficult due to the highly dynamic model.  Whereby, we hand-tuned the coefficients to achieve the necessary pleasing results.  We used simple convex shapes (i.e., boxes) to represent the character limbs as shown in Figure \ref{fig:biped_3d_parts}(c) and Figure \ref{fig:sim_2D_getup}.

\section{Experimental Results}
We applied our approach to different simulation situations to demonstrate the advantages of our method and its potential for creating rising motions without pre-recorded animation libraries (i.e., key-framed data).  The simulations were ran at 100 frames per second (fps) and were executed on an Intel Core i7-2600 CPU with 16-GB of memory running Windows-7 64-bit on a desktop PC.  The results are shown through a series of experiments to warrant the practicality and robustness of our approach for generating adaptive biped stepping motions without key-framed data.  In short, the visual results testify to the robustness and simplicity of our approach for synthesizing balancing get-up actions.  The overall computational time for generating the character motions with control, including dynamic simulation overheads (e.g., rigid body constraints and contacts) was on average less than 3 ms, respectively (i.e., better than real-time performance).

The controller generates essential information for the biped character to get-up.  This information pertains to the end-effectors' locations and upper body's posture.  With inverse kinematics and data-driven approaches, the generated motions are not physically accurate.  These approaches usually fail to produce realistic get-up poses, as the character's dimensions are changes, and do not reflect the strength of the character's muscles.  However, our simulations use torque and joint forces to move the final rigid body skeleton to an upright pose.

When it has been identified that the character has fallen down and has come to a complete stop.  Whereby, we wait for the character's angular and linear velocity to reach a minimum threshold, e.g., in-case he is sliding down stairs, or rolling.  Once the character has come to a complete stop, we engage the rise-up controller and monitor its progress at repeated intervals.  The generated rise motions where robust against initial posture poses (i.e., see Figure \ref{fig:fallen_pose_screenshots} and would roll to a comfortable (i.e., on their front or back) before moving their hand and feet contacts while shifting the CoM towards their feet.

Figure \ref{fig:sketchA} shows the model being applied to the sigattal plane to illustrate how we apply the base controller to our biped character model.  This can be compared with simulation results in 3D and 2D as shown in Figure \ref{fig:sim_getup} and Figure \ref{fig:sim_2D_getup}.

\figuremacroW
{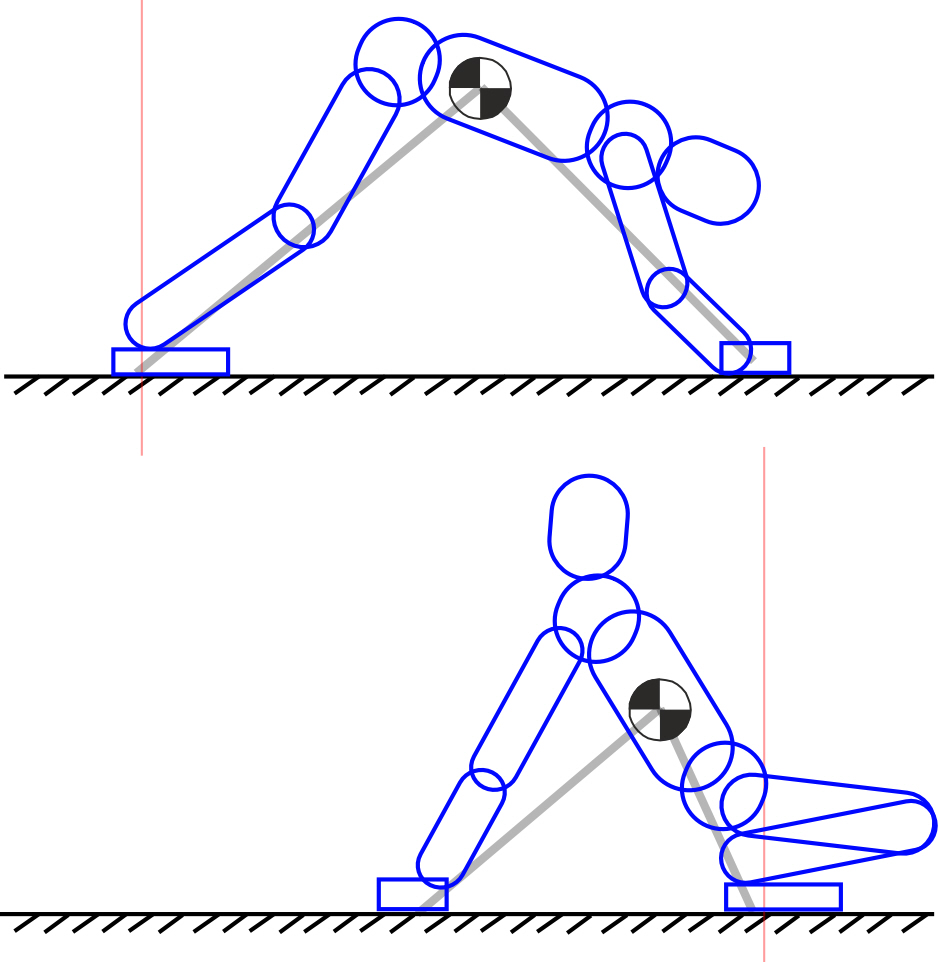}
{Low-Dimensional Model Controlling Articulated Skeleton}
{The key points when mapping from our low-dimensional controller model to a high dimension articulated character.  Our model can be applied to the character getting up from the front or the back.}
{0.7}

The preliminary work shows promising results, with a great deal of flexibility for improvement and adaptation.  Our simple model provides a robust, computationally fast, and controllable solution for generating fundamental balancing character pose information.

\figuremacroW
{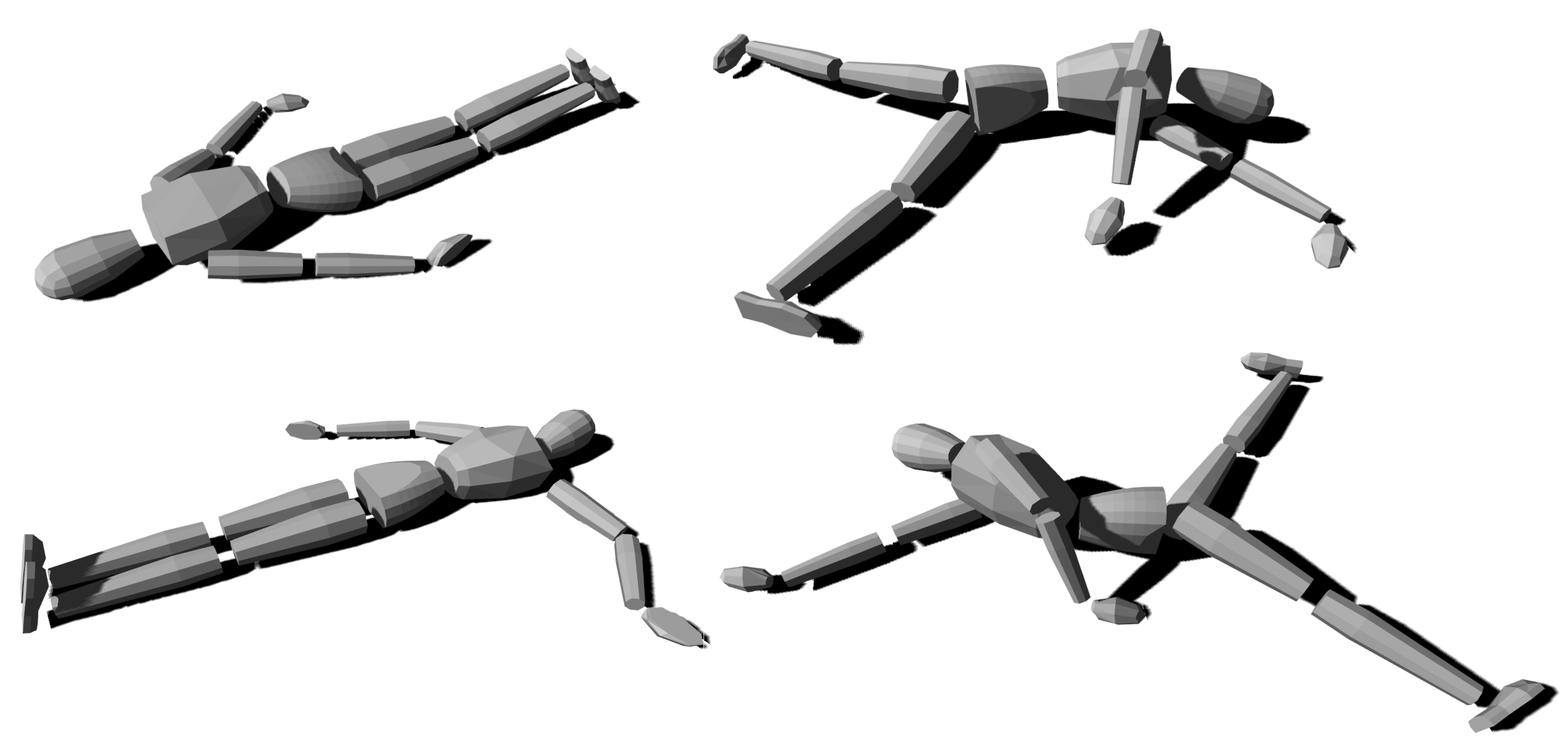}
{Simulation Snapshots}
{The rise controller was tested on a variety of starting poses.}
{1.0}

\section{Discussion}

\subsection{Limitations}
The simple base-model assumes a single-point mass and mass-less arms and legs.  Furthermore, our model focused on getting-up from either the front or back using both hands and feet, and does not address other approaches.  For example, rising tangentially (i.e., from the sides), using other contact points (e.g., the elbows and knees), and we do not include any feet or hand slipping.  Additionally, we only looked at static slow motions, whereby the projected ground CoM stayed within the support region to remain continuously balanced during movement transitions.  We did not address highly dynamic get-up motions where the character could gain momentum by swinging their body.  Although our get-up approach is been based on restrictive assumptions (i.e., from the front or back), the created motions proved to be visually plausible and life-like.

\section{Conclusion and Further Work}
We have presented a computationally simple, robust, and flexible approach for generating character rise animations.  We do not require any key-framed data (i.e., motion capture libraries) to generate the fundamental movements.  The biped character get-up motions are physically-plausible and balanced.  We enforce joint limits, non-sliding hand and feet contacts, and geometric environmental considerations.

Our experimental results show that our approach can be customized to produce a wide variety of basic rise motions (e.g., height of CoM, max/min, leg/arm extending, speed, front-back).  As the character's features are changed (i.e., support polygon and CoM) our model \textit{automatically} adapts the posture and contact placement information for the rise animation in real-time; for example, a designer would not need to compensate for any changes when creating the rise motion.  Furthermore, our approach can be combined with motion capture data (or random human-like rhythmic movements) to create more captivating and life-like motions that are more unique and possess a character's personality.

While the basic model has been introduced here, further work could be to investigate how the model copes with uneven terrain (e.g. on a slope).  Furthermore, when we are pushed over, we rotate and reach out in the direction we are falling.  Hence, we believe that our model can be adapted to other situations, for example, if the character loses its balance and is unable to recover, it could switch to the get-up motion logic so that the fall sequence is more natural looking (e.g., rotating in the direction of the fall and placing the arms out); compared with matching pre-canned animations to fit the unique fall situation which can look unnatural for that moment.


\section{Acknowledgments}

The author would like to thank the anonymous reviewers for taking time out of their busy schedules to provide helpful comments to make this a more concise, clear, and readable paper.


\bibliographystyle{acmsiggraph}
\bibliography{paper}


\figuremacroF
{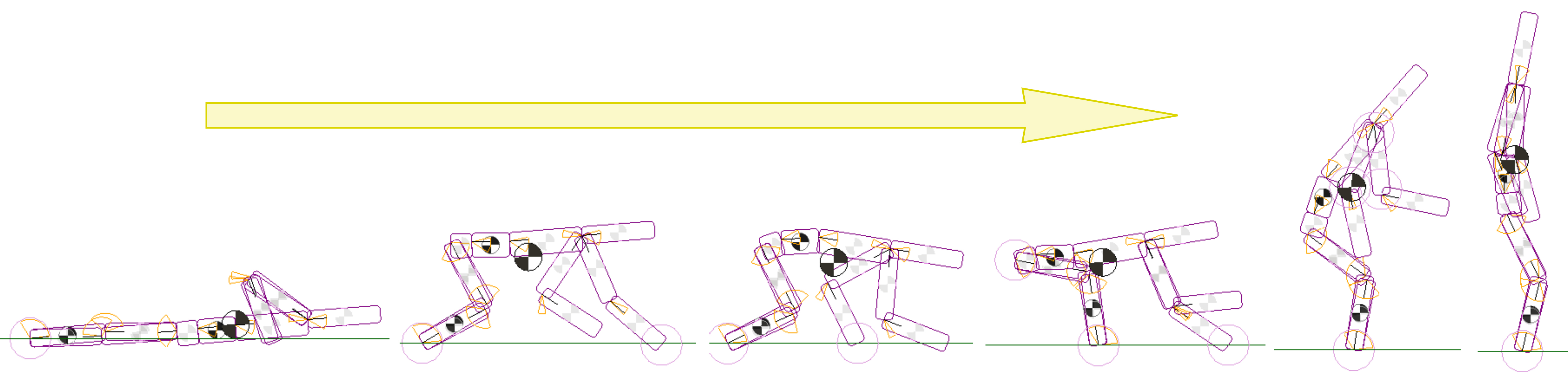}
{2D Rise Simulation}
{A simulation of a 12-link articulated 2D character to show the sequence of steps for our approach.  Initially, placing the hands and feet parallel to each other.  Release the contact for one hand.  Moving the CoM towards the feet while enforcing minimum and maximum distance constraint distances between the feet and hands.  Move free hand to new location.  Make free hand the support hand and release constraint for old support hand.  Repeat to move CoM above feet while maintaining distance constraints.  When CoM is above the feet, release hand constraints and rise upwards while keeping the CoM above the feet.  Finally, the character is standing upright and balanced.}
{1.0}

\figuremacroF
{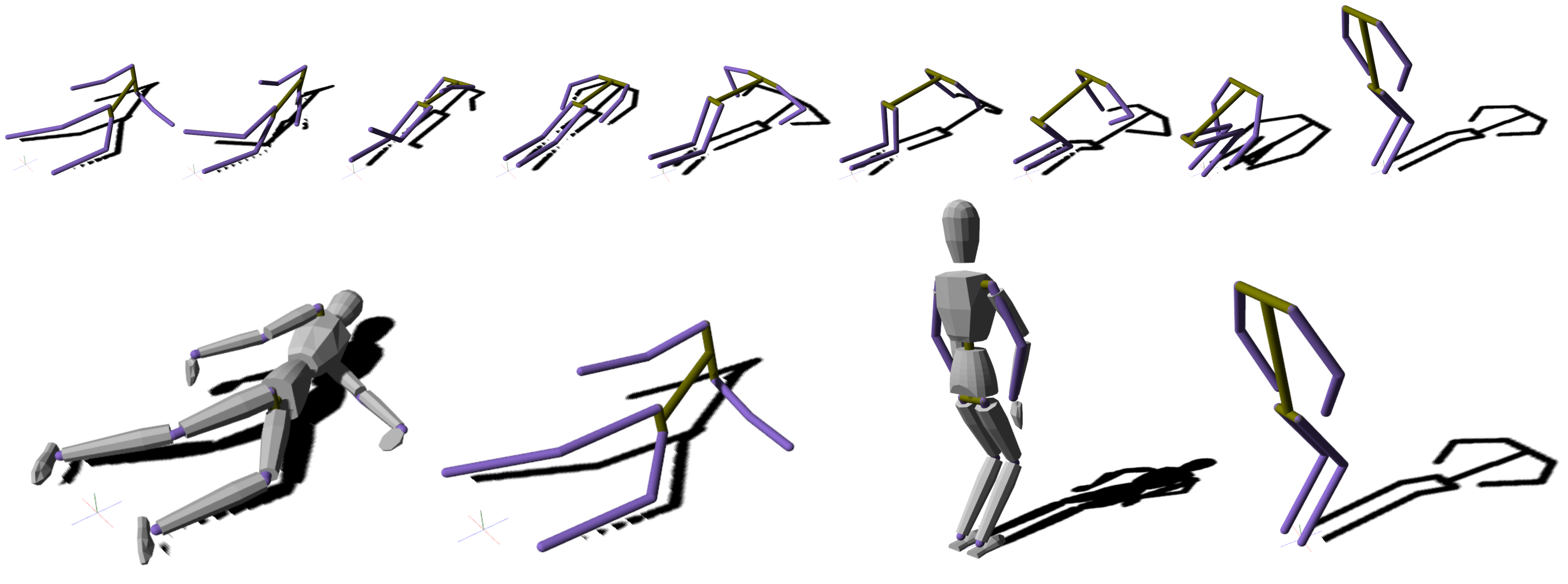}
{Rise Simulation Key Pose Snapshots}
{The rise controller model could be mapped onto skeletons of different levels of complexity to reduce ambiguity and singularity issues.  The figure shows the full-body skeleton represented as a basic 9-link stick man rolling over and getting-up from the front.}
{1.0}

\figuremacroF
{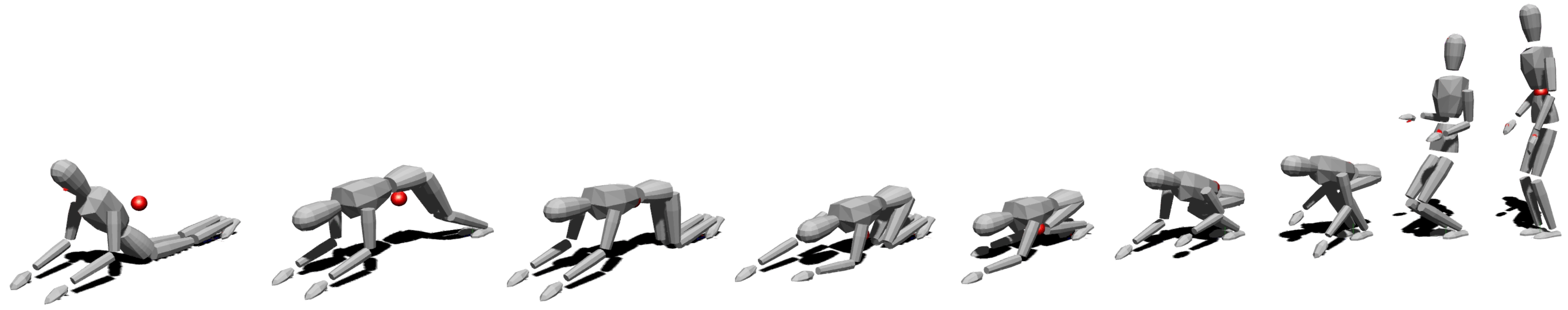}
{3D Rise Simulation}
{The character picks himself up by positioning himself on his from with his feet and arms to his left and right (i.e., not twisted or contorted), then proceeds to shift his center-of-mass towards his feet while iteratively being supported by his arms.  When the center-of-mass is above the feet support region the character can rise upwards to a full standing pose.}
{1.0}

\figuremacroW
{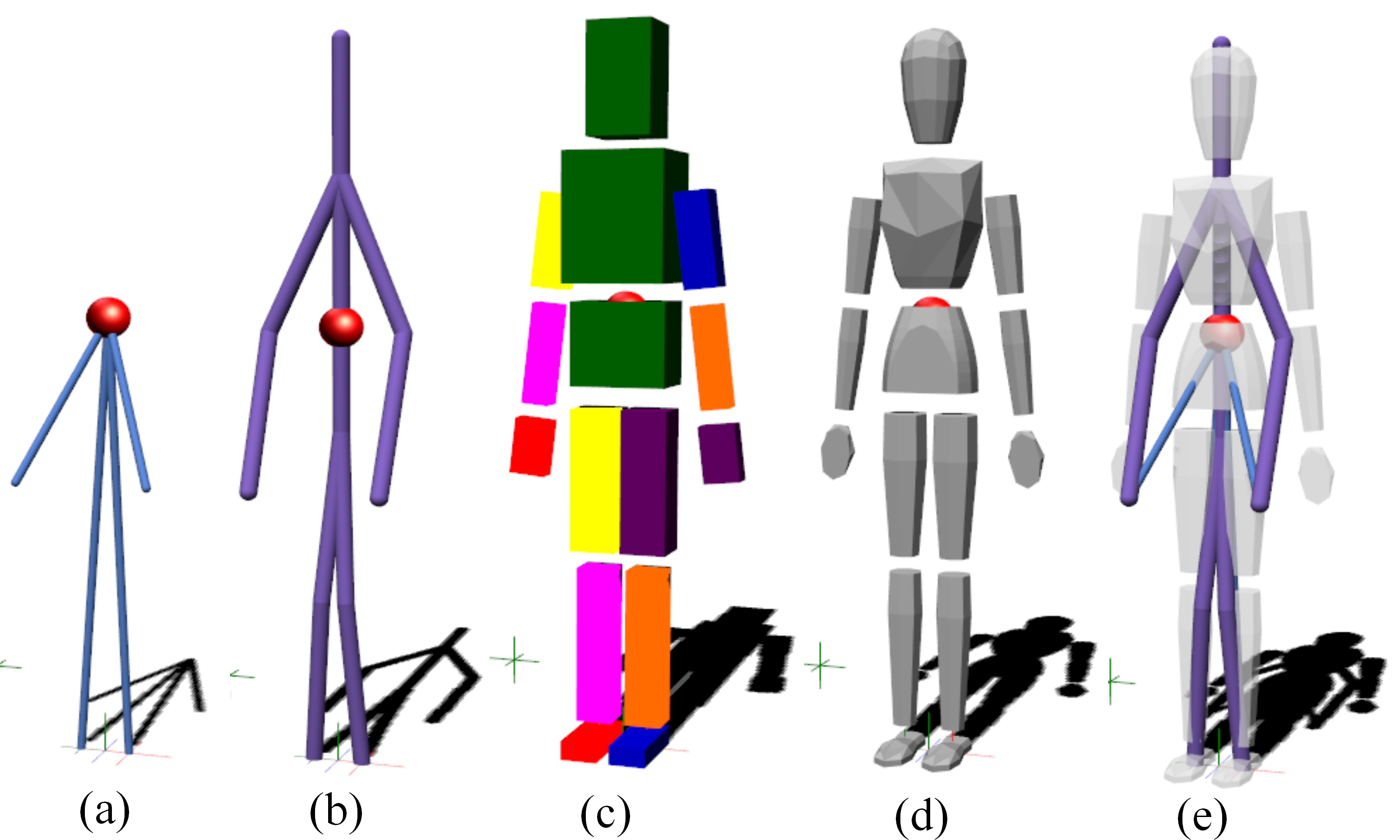}
{3D Biped Decomposition}
{Our approach for generating balanced biped get-up motions.  (a) Simple optimization model (i.e., CoM and end-effectors for hands and feet), (b) mapping onto an articulated structure, (c) rigid body bounding box approximation for contacts, (d) 15-link articulated 3D character, and (e) combined to formulate the full character solution.}
{1.0}

\figuremacroW
{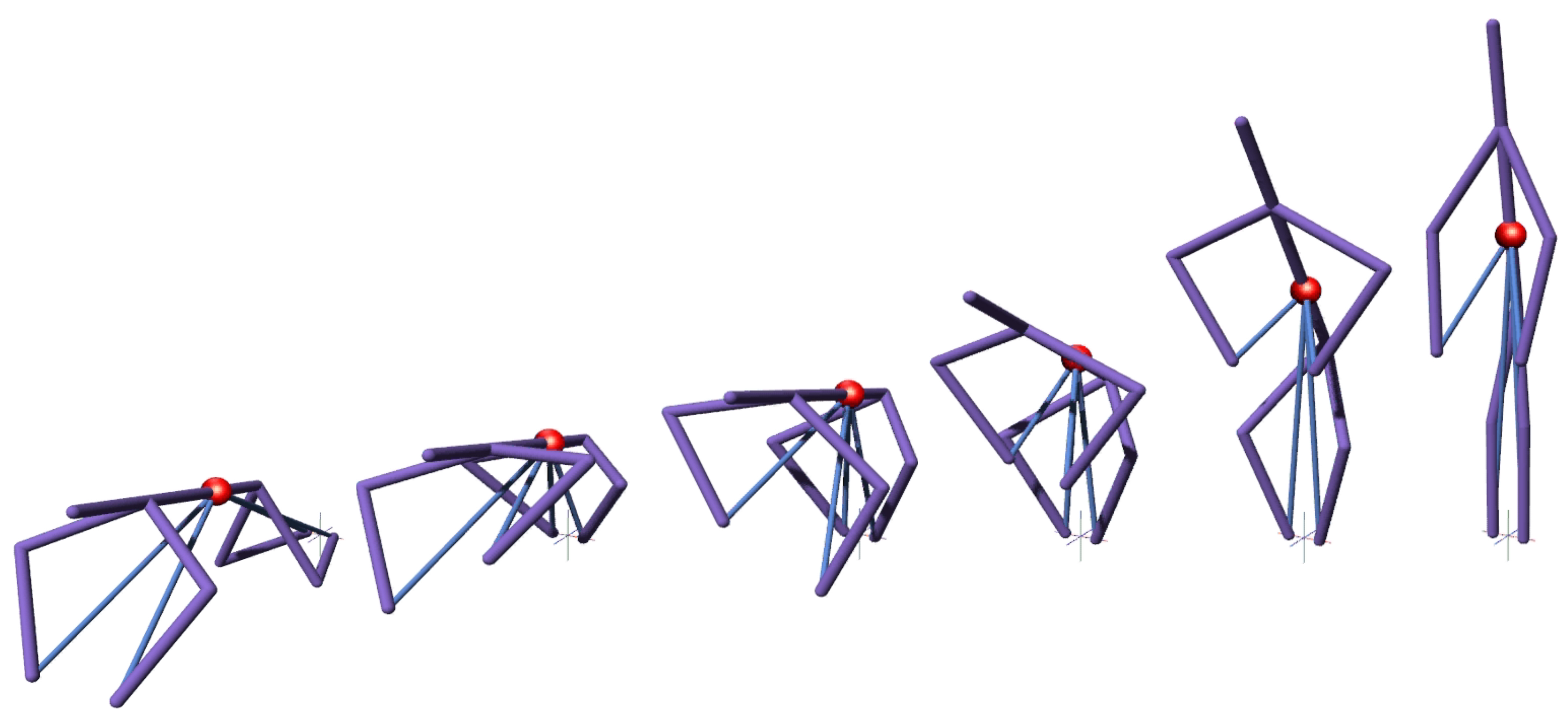}
{Simulation Snapshots}
{The optimized get-up model's solution mapped onto a 9-link articulated 3D skeleton.}
{1.0}

\figuremacroW
{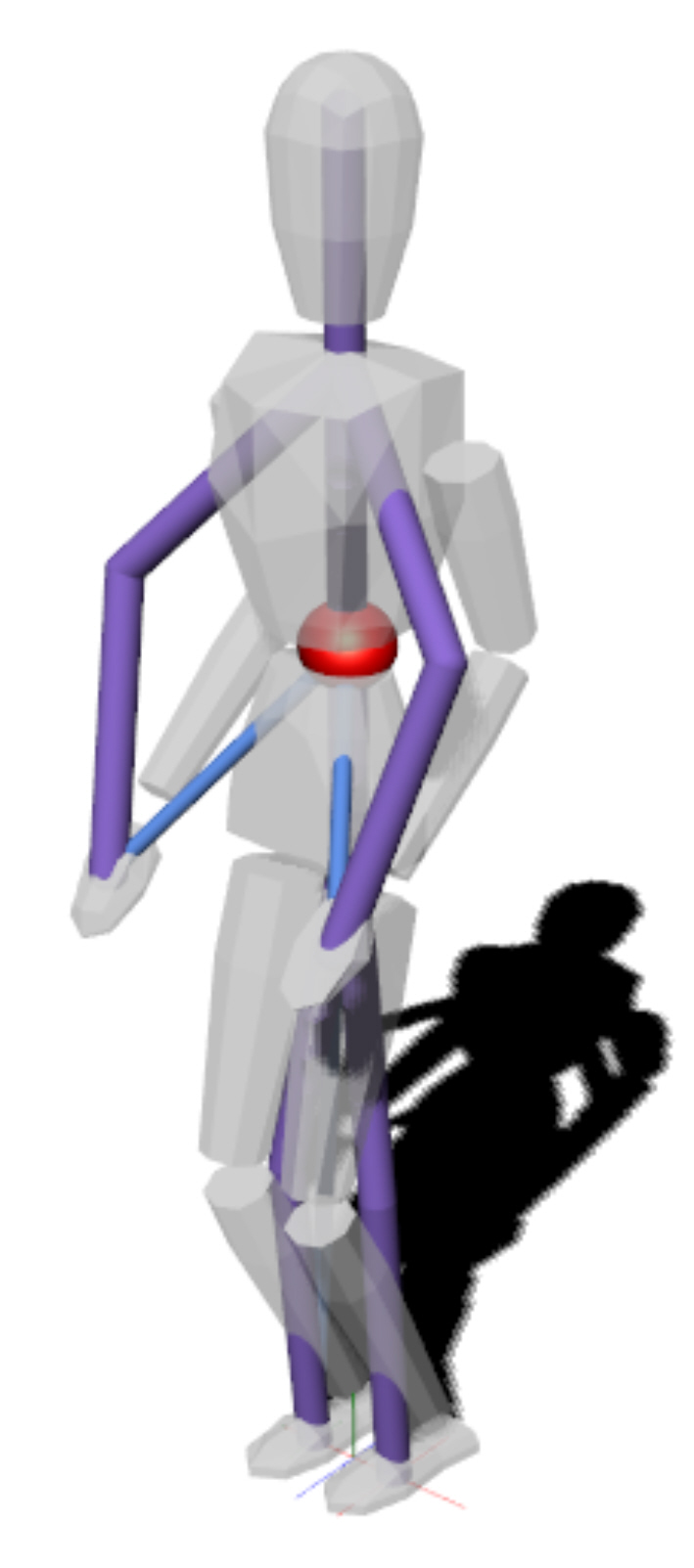}
{Biped Model}
{The optimized get-up model's solution mapped onto a 15-link articulated 3D skeleton.}
{0.4}

\end{document}